\documentclass[letterpaper, 10 pt, conference]{ieeeconf}  

\IEEEoverridecommandlockouts                              

\makeatletter
\let\NAT@parse\undefined
\makeatother
\usepackage[numbers, sort&compress]{natbib}

\usepackage{times}

\usepackage{tikz}
\usepackage{graphicx} 
\usepackage{amsmath} 

\usepackage{makecell}
\usepackage{kotex}

\usepackage{amssymb}  
\usepackage{amsfonts}
\usepackage{subfigure}
\usepackage{ctable}
\usepackage{makecell}

\usepackage[titlenumbered,linesnumbered,ruled,vlined]{algorithm2e}

\usepackage{fancyvrb}
\usepackage{xcolor}
\usepackage{verbatimbox}

\usepackage[labelsep=period]{caption}

\usepackage{multicol}

\usepackage{amsthm}
\theoremstyle{definition}

\usepackage{default_acronyms}
\usepackage{default_math}
\usepackage{default_SIunits}
\usepackage{default_misc}

\usepackage{soul,color}
\usepackage{verbatim} 
\usepackage{multirow}
\usepackage[bookmarks=true]{hyperref}
\usepackage{cleveref}

\usepackage{lipsum}
\usepackage{xcolor}

\DeclareMathOperator*{\argmin}{argmin}

\let\oldnl\nl
\newcommand{\nonl}{\renewcommand{\nl}{\let\nl\oldnl}}

\begin{document}
	
	\title{
		Low-cost Thermal Mapping for Concrete Heat Monitoring
	}
	
	\author{Alex Junho Lee${}^{1}$, Younggun Cho${}^{2}$ and Hyun Myung${}^{3*}, $~\IEEEmembership{Senior~Member, IEEE}%
		\thanks{$^{1}$Alex Junho Lee is with the Department of Civil and Environmental Engineering, KAIST, Daejeon, S. Korea {\tt\small alex\_jhlee@kaist.ac.kr}}
		\thanks{$^{2}$Y. Cho is with the Department of Electrical Engineering, Inha University, Incheon, S. Korea. {\tt\small yg.cho@inha.ac.kr}}%
		\thanks{$^{3}$Hyun Myung is with the School of Electrical Engineering, KAIST, Daejeon, S. Korea {\tt\small hmyung@kaist.ac.kr}}
		\thanks{$^{*}$Corresponding author: Prof. Hyun Myung}
	}
	
	\maketitle
	\IEEEpeerreviewmaketitle
	
	\acresetall
	
	\begin{abstract}

Robotics has been widely applied in smart construction for generating the digital twin or for autonomous inspection of construction sites. For example, for thermal inspection during concrete curing, continual monitoring of the concrete temperature is required to ensure concrete strength and to avoid cracks. However, buildings are typically too large to be monitored by installing fixed thermal cameras, and post-processing is required to compute the accumulated heat of each measurement point. Thus, by using an autonomous monitoring system with the capability of long-term thermal mapping at a large construction site, both cost-effectiveness and a precise safety margin of the curing period estimation can be acquired. Therefore, this study proposes a low-cost thermal mapping system consisting of a 2D range scanner attached to a consumer-level inertial measurement unit and a thermal camera for automated heat monitoring in construction using mobile robots.

\end{abstract}

	\section{Preliminaries}

In the construction industry, heavy equipment and human-robot interaction interfaces have been widely used to enhance efficiency and ensure the safety of construction sites. These technologies primarily focus on assisting and scaling up human manipulations, enabling faster and safer labor during construction procedures~\cite{saidi2016robotics}. 

The concept of smart construction has been introduced to provide greater autonomy as compared with assistive robots based on human manipulations. Smart construction topics cover the overall management of construction resources and elements (e.g., materials, equipment, and devices) for autonomy and interactivity~\cite{niu2016smart, myung2021robotic}. To achieve this goal, information from the construction site should be gathered and managed for delivery to the right person or system in proper format and provide a basis for precise decision-making~\cite{chen2015bridging}.

\begin{figure}[!t]
	\centering
	\includegraphics[width=\columnwidth]{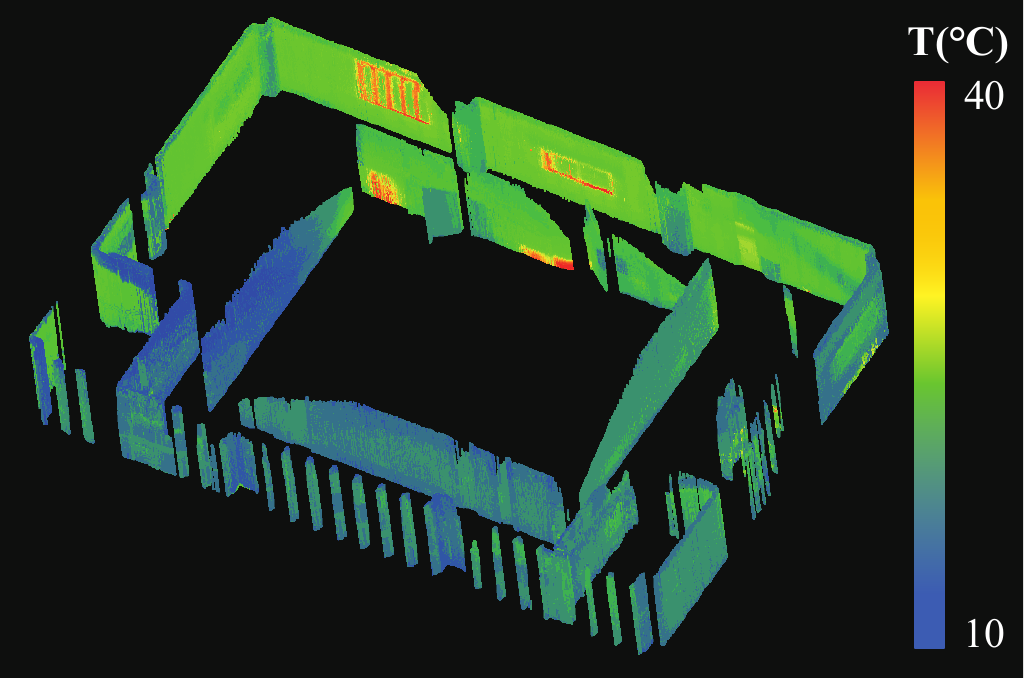}
	\caption
	{
	Sample of a thermal map reconstructed from our experiment. We used a low-cost mapping system with a range scanner, IMU, and thermal camera. We obtained the thermal point cloud map from SLAM using pose-graph optimization (PGO) by assuming a ground robot at a known height.
	}
	\label{fig:cover}
	\vspace{-4mm}
	
\end{figure}

However, smart construction has been only partially applied in the field due to application difficulties. For instance, a versatile perception algorithm and control mechanism are required to establish spatial interaction between a robot and the construction environment. A construction site continuously changes over time and on a large scale, and AI must be cognizant of up-to-date spatial information. Recently, simultaneous localization and mapping (SLAM) based on cameras~\cite{lim2014real,lim2022uv,song2022g2p} or light detection and ranging (LiDAR) systems~\cite{shan2020lio,lim2022quatro} have been introduced to help robots localize themselves and recognize spaces even in unseen or unknown environments. When these localization and mapping capabilities are combined with path planning and automated inspection technologies, the structural information of a site can be periodically updated, and a construction project can be monitored in real time.

The information updated by an autonomous system could include 3D structures, tracking results of dynamic objects, geodetic surveying, etc. In our study, we concentrated on the thermal behavior of a construction site with respect to both the safety and energy efficiency of buildings. When concrete is cured, the accumulated heat over time~\cite{bergstrom1953curing} is a critical variable for calculating the required curing period, and the temperature change rate should remain within a certain range~\cite{kim1998compressive}. Thus, at construction sites, adjustments to the curing period or maintenance of the temperature of the concrete based on the atmosphere is necessary. However, criteria based on air temperature provide an indirect estimation of the accumulated heat and make it possible for defects or accidents to occur. As an alternative approach, we propose that the concrete's surface temperature be used to calculate the accumulated heat by providing a SLAM-generated thermal point cloud from a low-cost system of sensors mounted on a ground robot as in Fig. \ref{fig:cover}. Assuming concrete walls of known height in a curing period, we show that a thermal point cloud of the walls can be acquired using only a 2D range scanner combined with a consumer-level IMU and a thermal camera. Throughout this study, we introduce a method for building thermal point clouds of walls using the proposed sensor system and compare the mapping results derived from experiments.

\begin{figure*}[!t]
	\centering
	\includegraphics[width=0.85\textwidth]{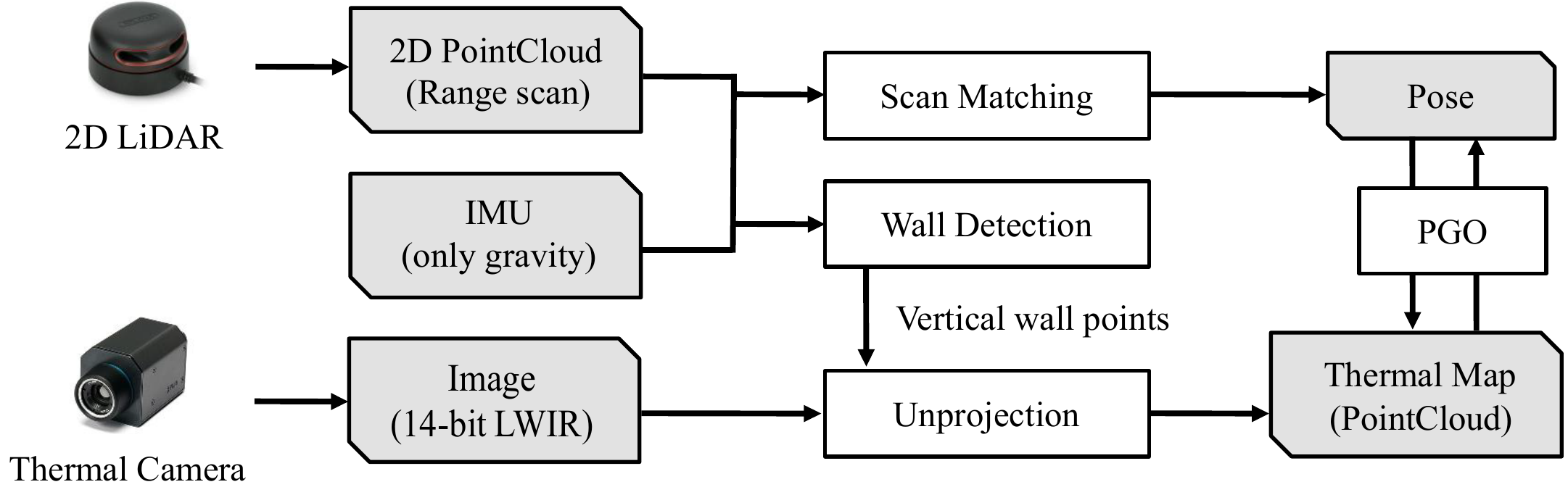}
	\vspace{1mm}
	\caption
	{
		Outline of our algorithm. We used a sensor system consisting of a 2D LiDAR combined with a consumer-level IMU and a thermal camera. We used only the gravity direction from an IMU to compensate for the range scans and used the compensated range scans for scan matching and wall detection. After the vertical walls were detected, the points were projected onto thermal images to obtain temperature information and then unprojected to a point cloud. Then, with the constructed point cloud map and pose information from scan matching, we refined the final pose with the map by PGO.
	}
	\vspace{-3mm}
	\label{fig:outline}
\end{figure*}

	\section{Thermal-LiDAR SLAM}

A robot pose can be described in 6 variables ($x$, $y$, $z$, roll $\theta_{x}$, pitch $\theta_{y}$, and yaw $\theta_{z}$).
However, because we rely on a 2D range scanner to estimate the depth of a scene, we constrained the robot motion in 2D space ($x$, $y$, $\theta_{z}$) for initial pose estimation. To obtain the pose and map simultaneously, we used SLAM based on 2D range scans, as shown in the pipeline in \figref{fig:outline}. We projected the range scans into a horizontal plane using the gravity direction detected from the IMU and used scan matching to estimate the relative transformation between scans. Then, assuming that a mobile robot is operating at a fixed elevation in a building layout with a known floor height, we generated vertical wall points and assigned the corresponding thermal values. Then, through unprojection, we obtained the thermal map as a point cloud, and the relative transformations between its poses were refined by pose-graph optimization (PGO) based on both the geometry and thermal values of the wall point cloud.

\subsection{Scan Matching and Pose Estimation}

To identify the initial poses, we first projected the range scans from a 2D LiDAR into the $xy$ plane using the gravity vector $g$ obtained from the IMU. Because the update rate of the IMU ($\sim$ 100 Hz) was higher than that of the range scanner ($\sim$ 10 Hz), we searched for the IMU message with the closest timestamp to each scan message and assigned the gravity vector to points $\text{x}_{\textsf{S}}$ in each scan $\textsf{S}$. Assuming that a robot seldom experience a large slope ($\theta_x$,~$\theta_y \simeq$~0), the laser scans mostly fall on the wall surface. Using the scanned wall points, we next calculated the projection of the scan points toward an imaginary plane at the level of the mobile robot and gravity vector $g$ as a plane vector: $\textsf{x}_{xy}~=~\text{x}_{\textsf{S}}~-~\cfrac{\text{x}_{\textsf{S}}~\cdot g}{g \cdot g}~g$. Using the projected scan points in 2D, $\textsf{x}_{xy}$, we then calculated the relative transformation $\textsf{T}_{ij} (x, y, \theta_z)$ between scans of nodes $i$ and $j$, $l_i$ and $l_j$ to minimize the reprojection error~$\mathsf{e}$:
\begin{equation}
	\argmin_{\textsf{T}_{ij}} \mathsf{e} = ||l_i - \textsf{T}_{ij}(l_j)||.
\end{equation}

\begin{figure*}[!t]
	\centering
	\includegraphics[width=\textwidth]{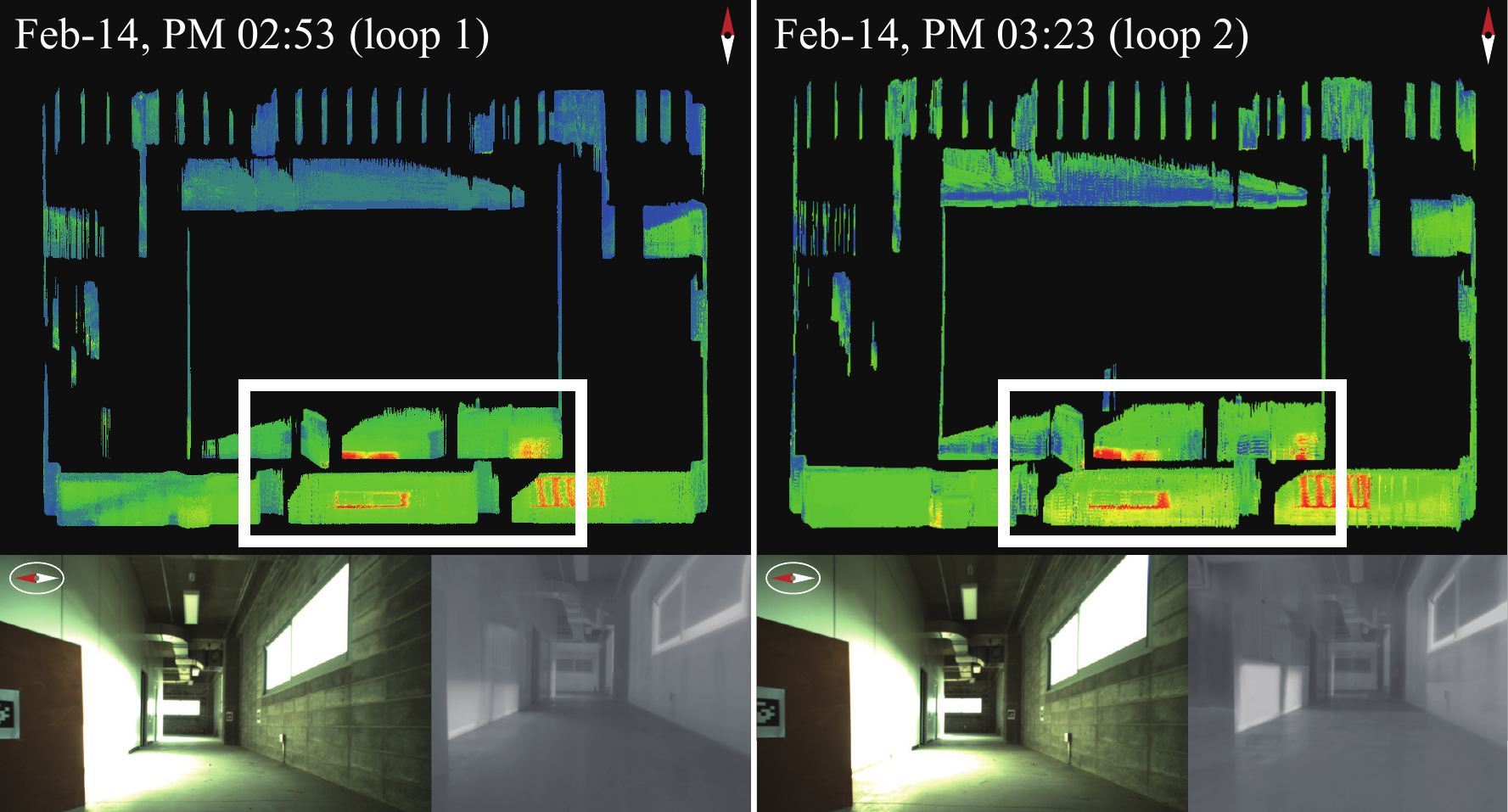}
	
	\caption
	{
Examples of constructed thermal maps with RGB and thermal images obtained at each space. In both runs, walls facing sunlight showed higher temperature (green) than opposing walls (blue). In addition, due to the direct sunlight, window frames exhibited higher temperatures. This can also be seen in the thermal image displayed at the lower right.
	}
	\label{fig:sample}
\end{figure*}

\subsection{Thermal Map Generation}

After acquiring the relative transformations between the scans, we generated the thermal point clouds. From the 2D-projected laser scans at the level of the mobile robot, we expanded the scan points at the elevation of a robot~$\textsf{x}_{xy}$ vertically by repeating points in the gravity direction~$g$ and opposite the gravity direction~$-g$. Because we assumed that the floor height and sensor elevation were known, we could easily calculate the wall points~$\text{p}_{wall}$ around the robot.

Then, the vertical wall points~$\text{p}_{wall}$ are projected onto the thermal image $I$ with known intrinsic matrix $K$ and extrinsic transformation $\text{T}$. By transforming the wall points into thermal image coordinates $\text{u} = (u, v)$ using $\text{u}~=~K\cdot (\text{T}\cdot \text{p}_{wall})$, we obtained the image coordinates of each wall point. In this procedure, points outside the thermal camera's field of view were filtered out. Then, with the thermal values in the image $I(\text{u})$ assigned to each corresponding point, the temperature $\mathcal{T}$ from the thermal image was saved in the intensity field of the thermal point cloud $\mathbf{x}$.

While we obtained the whole poses and thermal point clouds assigned to each node, we ran PGO based on the geometry of points and intensity. We used the Ceres solver~\cite{agarwalceres2022} in this procedure, setting up the loss to optimize both the geometry and temperature difference. Between selected point pairs $\mathbf{x}_i$ and $\mathbf{x}_j$ of nodes $i$ and $j$, we calculated the optimal transformations $\text{T}$ between nodes to minimize the error $\text{e}$:
\begin{equation}
	\argmin_{\text{T}}\text{e} = \sum_{(i,j)}(||\mathcal{T(\mathbf{x}}_i) - \mathcal{T(\mathbf{x}}_j)||
	+ ||\mathbf{x}_i - \text{T}_{ij} \cdot \mathbf{x}_j||).
\end{equation}
We calculated the optimal relative transformation between nodes by applying this error function for every selected point pair between the selected nodes. We then obtained the optimized poses and their corresponding thermal point clouds $\mathbf{x}$. We accumulated the thermal point clouds of every node given the poses of the nodes and obtained the final thermal point cloud.

	\section{Experiments and Results}

We run our experiments on a construction site at Heunghae-eup, Pohang, South Korea, in Feb. 2020, in which a concrete building was part of the actual construction. We used a sensor system that was mounted on a mobile robot, and we recorded the data in a rosbag file. To test the sensor reliability under the environmental variances of different construction sites, we executed multiple repetitions at different times and based on different noise sources such as haze. We plan to release the data used in the experiment in the near future.

\subsection{SLAM}
\begin{figure}[!b]
	\centering
	\includegraphics[width=\columnwidth]{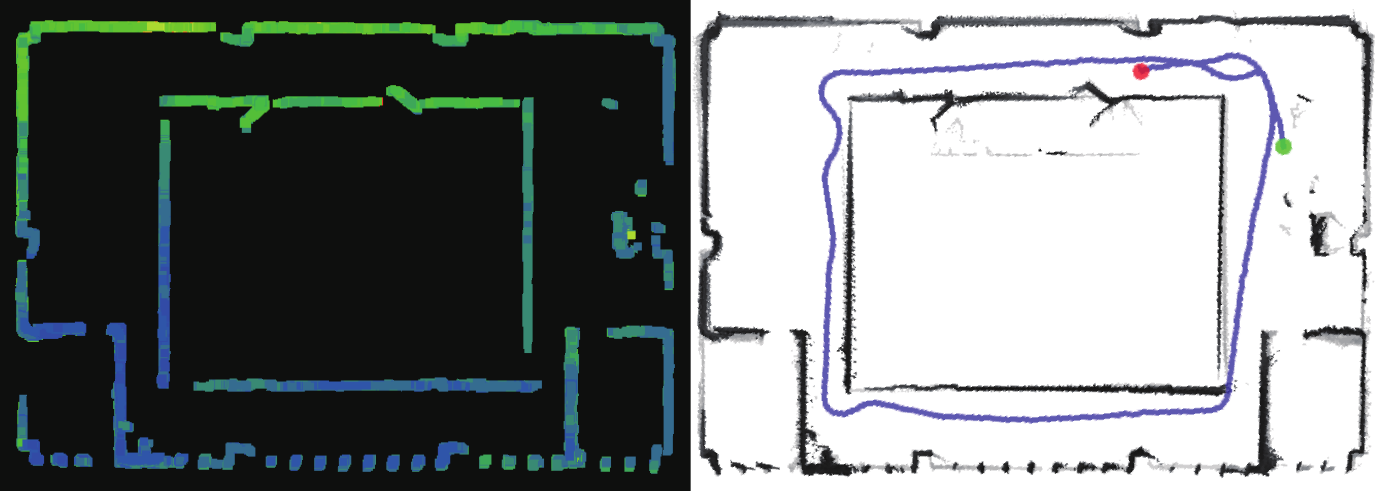}
	\caption
	{
Generated maps from thermal point clouds at loop~1~(left) and 2D SLAM at loop~2~(right). The green and red dots on the right are the start and end points of the estimated SLAM trajectory. With the wall layouts obtained from the generated map, we could monitor how precisely the walls were located during construction.
	}
	\label{fig:reconresult}
\end{figure}

During optimization, we used a levenberg-marquardt (LM)~\cite{marquardt1963algorithm} based solver combined with the huber loss~\cite{huber1992robust}. To ensure robust optimization, the translation and rotation weights were set to 5.0 and 400.0. In \figref{fig:sample}, we present the results of our method using our dataset. The obtained thermal maps and sample images from each run were plotted. Although our algorithm assumes that no other type of structure exists other than plain walls in the environment, the results of SLAM succeeded in estimating the pose and in constructing an aligned point-cloud map. In \figref{fig:reconresult}, we compare the thermal map and results of 2D SLAM based on Google Cartographer~\cite{hess2016real} before temperature-based PGO using LM solver in Ceres, from different runs, respectively. As the figure shows, consistent layouts were reconstructed and our SLAM pipeline did not diverge or become lost during the two experiments.

\subsection{Thermal Mapping}
\begin{figure}[!t]
	\centering
	\vspace{2mm}
	\includegraphics[width=0.9\columnwidth]{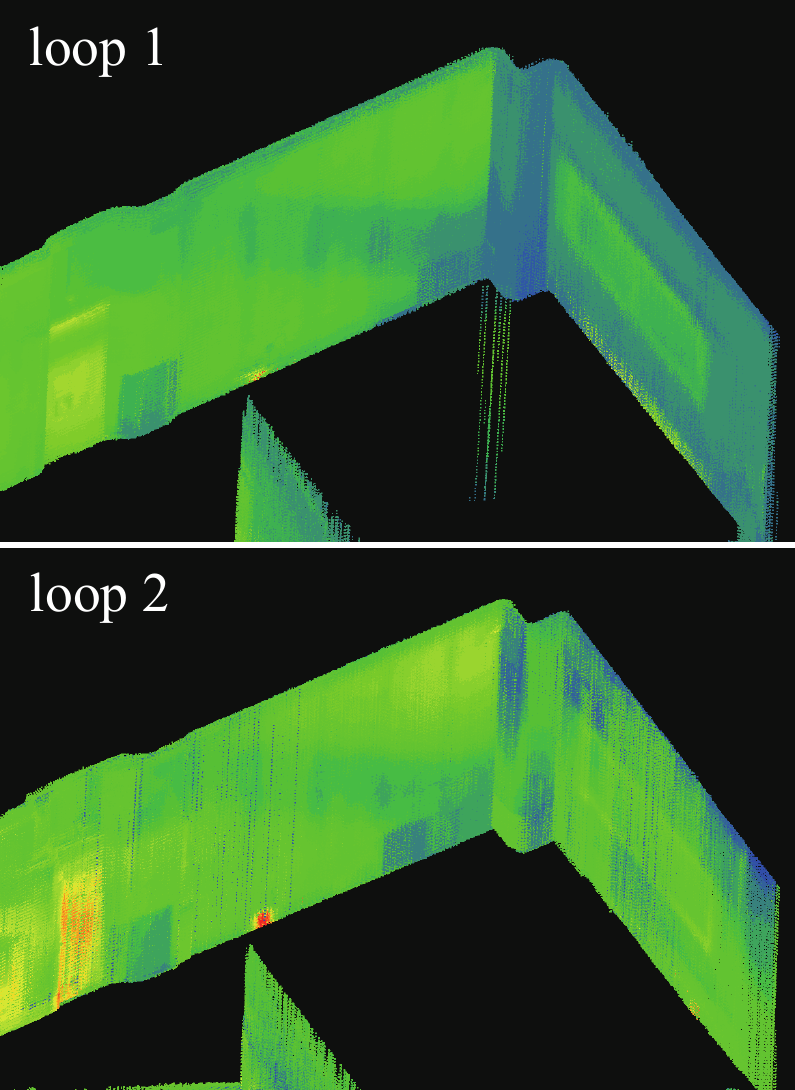}
	\vspace{2mm}
	\caption
	{
Measured temperature of the same corner wall at different times after alignment. Through ICP alignment, thermal point clouds were easily registered, and measurement points are successfully monitored over time.
	}
	\label{fig:therresult}
\end{figure}

Using the algorithm suggested in the previous section, we built a thermal point cloud of the environment through two iterations as derived from PGO. As shown in the upper parts of \figref{fig:sample}, clear wall point clouds were extracted, where the temperature of each wall is shown in a rainbow color scale from 10$^\circ$C to 40$^\circ$C. In \figref{fig:sample}, the point clouds are rotated to face north. As the experimental site was in the northern hemisphere of the Earth, heat dissipation from the Sun was more concentrated on the southern (lower) walls of the structure during the daytime. As a result, a smaller blue area (lower temperature) on the walls facing south could be observed. In addition, we observed that the overall temperature of the construction site increased as time passed based on temperature visualization. The temperature difference could be easily monitored by running a simple registration algorithm such as the iterative closest point (ICP)~\cite{besl1992method} between thermal point clouds from experiments, as shown in \figref{fig:therresult}.

	\section{Conclusion and Future Works}

In this study, we proposed a low-cost thermal mapping system for automated thermal monitoring during concrete curing. We also proposed a 2D range scanner system combined with an IMU and thermal camera to estimate the thermal point clouds of concrete walls. For more precision, we also suggested thermal-based PGO to optimize the obtained pose and point cloud. We hope our study can be used as a reference in ensuring that the temperature constraint in concrete curing is maintained and in helping to prevent accidents in construction due to inappropriate estimation of the accumulated heat in concrete.

Our method does have limitations. Specifically, it uses only vertically compensated 2D scans for vertical wall detection and cannot distinguish between walls and non-walls such as handrails or obstacles. Thus, in future studies, non-wall objects should be filtered by combining them with the additional constraints from thermal images. This will enable our work to be expanded from 2.5D (only vertical planes) to 3D.

	\renewcommand*{\bibfont}{\small}
	\bibliographystyle{IEEEtranN} 
	\bibliography{string-short,references}
	
	
	\vfill
	
\end{document}